%% file: main.tex
\pdfoutput=1
\documentclass[letterpaper, 10 pt, conference]{ieeeconf}
\IEEEoverridecommandlockouts   %
\usepackage{amsmath,amssymb}

\usepackage{amsthm}
\newtheoremstyle{tight}{2pt}{2pt}{\itshape}{}{\bfseries}{.}{ }{}
\theoremstyle{tight}
\newtheorem{assumption}{Assumption}
\newtheorem{proposition}{Proposition}
\usepackage{graphicx}
\usepackage{booktabs}
\usepackage{array}
\usepackage[table]{xcolor}
\usepackage{tikz}  %
\usetikzlibrary{arrows.meta,calc}

\definecolor{tabaccent}{HTML}{1A5FB4}  %
\definecolor{tabdark}{HTML}{0F3B6E}    %
\definecolor{tabband}{HTML}{CFE0F5}    %
\definecolor{tabink}{HTML}{111111}
\definecolor{tabmuted}{HTML}{454C55}   %
\newcommand{\thd}[1]{\textcolor{tabink}{\bfseries #1}}
\newcommand{\grouprow}[2]{%
  \multicolumn{#1}{@{}l}{\itshape #2}\\}
\newcommand{\mc}[1]{\multicolumn{1}{c}{#1}}
\newcommand{\unit}[1]{{\color{tabmuted}#1}}

\usepackage{microtype}
\newcommand{\best}[1]{\textcolor{tabaccent}{\bfseries #1}}

\usepackage{cite}
\usepackage{url}
\usepackage{flafter}
\definecolor{linkgreen}{HTML}{1E7B4F}  %
\usepackage[colorlinks=true,linkcolor=linkgreen,citecolor=linkgreen,urlcolor=linkgreen]{hyperref}
\makeatletter
\def\section{\@startsection{section}{1}{\z@}{2.5ex plus 0.5ex minus 0.5ex}%
{0.7ex plus 0.3ex minus 0ex}{\normalfont\normalsize\centering\scshape}}%
\def\subsection{\@startsection{subsection}{2}{\z@}{2.5ex plus 0.5ex minus 0.5ex}%
{0.7ex plus 0.3ex minus 0ex}{\normalfont\normalsize\itshape}}%
\makeatother

\graphicspath{{./}{assets/}}

\begin{document}

\title{\LARGE \bf SCRUB: Contact-Based Surface-Frame Estimation for
Force--Motion Control with an Armless Quadruped}

\author{Sourav Selvaraj, Juan Pablo Espinosa Chessal, and Mahdi Agheli%
\thanks{This work has been submitted to the IEEE for possible publication.
Copyright may be transferred without notice, after which this version may
no longer be accessible.}%
\thanks{The authors are with the Department of Robotics Engineering,
Worcester Polytechnic Institute, Worcester, MA, USA.
Corresponding author: Mahdi Agheli (e-mail: mmaghelih@wpi.edu).}}

\maketitle
\thispagestyle{empty}
\pagestyle{empty}
\raggedbottom

\begin{abstract}
Wiping, sanding, and inspection robots press a sliding tool on a surface with
controlled force.  Controlling that force needs the surface normal,
which is unknown on real-world, unmapped surfaces.  Surface-normal estimators can help estimate the surface normal, but they either need a friction model or assume that the instantaneous tool velocity lies along the surface; however, walking violates this assumption.
We propose SCRUB, a contact-based surface-frame estimator, which lets an armless quadruped slide a
body-mounted tool along near-vertical surfaces in the absence of known, axis-aligned, or vision-predicted normals. SCRUB's predictive
controller presses and slides the tool using a Kalman-filter estimate of
the normal from two cues: 1) A force cue measures the estimate's tilt
across the motion.  If friction directly opposes the motion, the cue is
zero exactly when that tilt is zero, for any friction coefficient.
However, if the cue's sliding direction is wrong, friction appears as a
false tilt.  2) A velocity cue measures tilt along the motion.  Walking
and pressing move the tool in and out.  SCRUB therefore subtracts the commanded
press rate from the tool velocity, averages the result over each gait
cycle, and treats this mean as lying along the surface.  On
Unitree Go1 hardware at nominal speed and load, SCRUB completed every
wall traverse and column orbit.  No orbit completed with
either cue removed, without gait averaging, or with a fixed normal.
\end{abstract}

\begin{keywords}
Legged Robots, Force Control, Contact Modeling,
Multi-Contact Whole-Body Motion Planning and Control, Force and
Tactile Sensing.
\end{keywords}
\begin{figure}[!t]
 \centering
 \includegraphics[width=0.52\columnwidth]{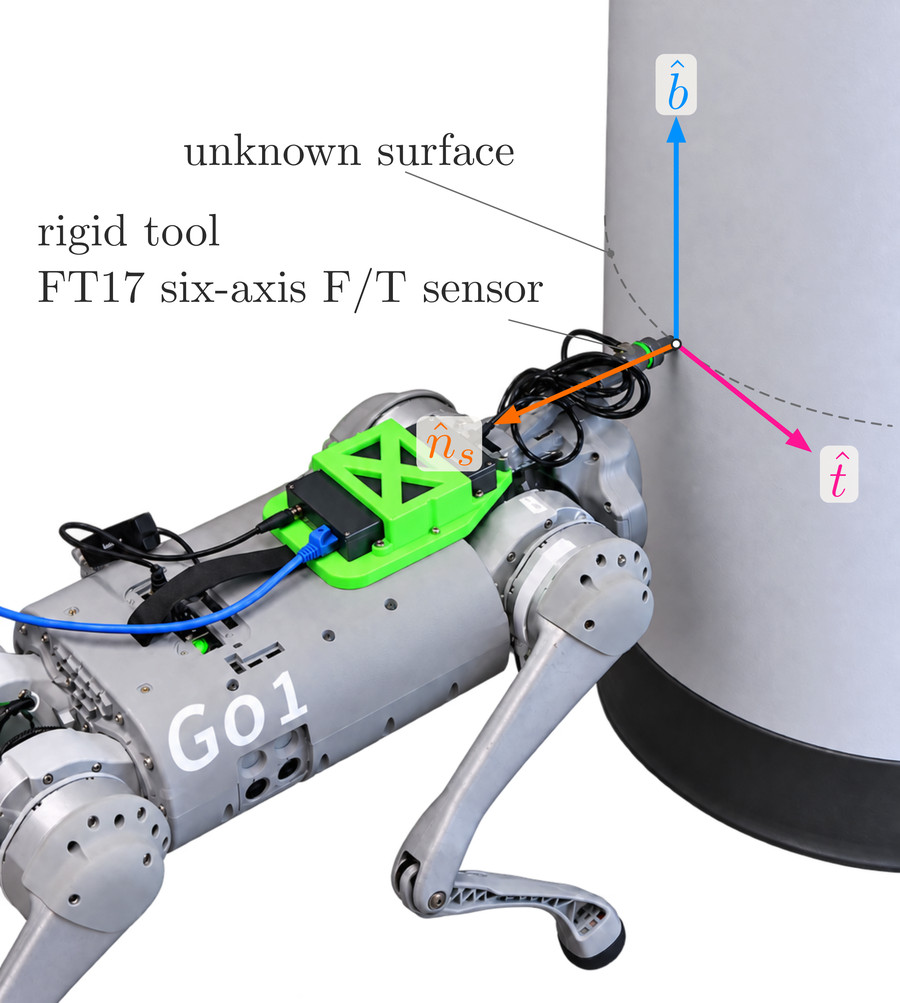}
 \vspace{-0.8ex}
 \caption{\textbf{Hardware and surface frame.}  Go1 with the
 trunk-mounted tool (rigid spherical tip in a silicone bumper), inline FT17
 force/torque sensor, and surface frame $(\hat t,\hat b,\hat n_s)$:
 tangent along the slide, binormal across it in the surface (near
 vertical here), and the adopted outward normal used by the predictive and
 whole-body controllers.  Background removed.}
 \label{fig:go1_axes}
 \vspace{-1.0ex}
\end{figure}

\input{sections/01_introduction}
\input{sections/03_setup}

\input{sections/04_estimation}
\input{sections/05_control}
\input{sections/06_results}

\input{sections/07_conclusion}

\section*{Acknowledgment}
\looseness=-1 The authors used Claude (Anthropic) to 
compute summary statistics, write the plotting code, check references, and help debug the robot software.

\bibliographystyle{IEEEtran}
\bibliography{references}

\end{document}

%% file: sections/01_introduction.tex
\section{Introduction}
\label{sec:intro}
\input{sections/fig_architecture}

\looseness=-1 Robotic wiping and sanding need force--motion control: the robot presses a tool along the surface normal while sliding it along the surface~\cite{raibert1981hybrid,khatib1987unified,mason1981compliance}. The controller therefore needs the local surface frame: the normal and two directions in the surface~\cite{villani2016force}. We study force--motion control on an armless quadruped that slides a trunk-mounted tool along a flat wall and around a curved column (Fig.~\ref{fig:go1_axes}). Without an arm, the robot saves mass and cost, but its trunk and stance feet must move the tool, resist the contact force, and keep the robot balanced~\cite{orin2013centroidal,sleiman2021unified}. Hence, a frame error makes the tool slide in the wrong direction and changes the contact wrench the legs must balance.

\looseness=-1 On an unmapped curved surface like the column of Fig.~\ref{fig:go1_axes}, the frame is unknown and rotates as the tool slides, and deformation of the tool's soft silicone bumper shifts the effective normal~\cite{johnson1985contact}. Since the bumper also hides the contact patch from vision, we estimate the frame from contact. Contact-based estimation faces two problems. First, friction tilts the force off the normal, and correcting that tilt needs a friction model~\cite{johnson1985contact,bicchi1993contact}. Second, the controller moves the tool in and out to hold the force, and the gait compresses and releases the bumper (\emph{gait ripple}). The tool velocity is therefore not always tangent.

\looseness=-1 Hybrid and operational-space controllers take the frame as given~\cite{raibert1981hybrid,khatib1987unified}, whereas constraint-based task specification estimates uncertain geometry within a known task model~\cite{deschutter2007constraint}. Estimators recover constraints from force~\cite{bicchi1993contact}, from force and motion on robot arms~\cite{karayiannidis2009adaptive}, or by touch with a leg~\cite{lin2020contact}. Three others need no known friction coefficient. One takes the normal perpendicular to the tool velocity~\cite{karayiannidis2014online}; another places it in the plane of the contact force and sliding direction~\cite{focchi2018slip}; and a third subtracts friction drag with a coefficient fitted online~\cite{nasiri2024hybrid}. However, each assumes that the instantaneous tool velocity is tangent, an assumption that is violated with walking gait.

\looseness=-1 Legged robots learn end-effector force control~\cite{portela2024force} and plan or learn interaction forces with an arm~\cite{bellicoso2019alma}, a leg~\cite{arm2024pedipulate}, multiple contacts~\cite{sleiman2023multicontact}, or the body~\cite{rigo2023contact}. Mahmud et al.~\cite{mahmud2026contact} slide a body-mounted structure, as extra support for an arm, on planes with known normals. Zeng et al.~\cite{zeng2026force} regulate wiping force on a wheeled-legged robot with the normal aligned to a controller axis, and humanoids wipe walls in a calibrated frame~\cite{samadi2021humanoid} or along normals predicted from vision~\cite{shi2026minimalist}. Focchi et al.~\cite{focchi2018slip} estimate the normal under a sliding foot, but none of these robots estimates the surface frame from contact for a sliding tool.

\looseness=-1 To estimate the rotating frame without a surface map, we propose Surface Contact Regulation Using the Body (SCRUB) as shown in Fig.~\ref{fig:architecture}. A Kalman filter corrects its normal estimate using two cues, called \emph{measurement rows}, including force and velocity rows (Fig.~\ref{fig:cues}), which measure the estimate's tilt across and along the slide. If friction acts exactly opposite to the motion, the force lies in the \emph{sliding plane} of the true normal and sliding direction~\cite{focchi2018slip}. The \emph{force row} therefore measures cross-slide tilt from the force component perpendicular to the estimated sliding plane. As along-slide tilt barely affects the force row, the \emph{velocity row} measures it from the tool velocity~\cite{karayiannidis2014online,focchi2018slip}. This row averages the tool velocity minus the commanded press rate over each gait cycle. Since most gait ripple cancels in this mean, the row treats the mean as tangent. Using the estimate, centroidal model predictive control (MPC)~\cite{orin2013centroidal,farshidian2020ocs2,leggedcontrol} plans body motion, foot forces, and the tool reaction force. Whole-body control (WBC) turns the plan into joint torques~\cite{zapolsky2014inverse}. Our contributions are:
\begin{enumerate}
\item A contact-based surface-frame estimator for a walking robot's sliding tool,
  with a gait-averaged velocity row and a force row that extends the
  sliding-plane cue of~\cite{focchi2018slip} (Secs.~\ref{subsec:normal_estimation}--\ref{subsec:curvature_estimation}). Ideally, the force row is zero exactly when the cross-slide tilt is zero, for any friction coefficient and even when the tool velocity has a normal component (Proposition~\ref{prop:rf}).
\item An interaction-wrench
  MPC~\cite{sleiman2021unified,sleiman2023multicontact,minniti2021model}
  adapted to an armless quadruped's trunk-mounted tool. The MPC plans the tool
  reaction with a first-order lag, a drag cost, and a cap on the binormal force
  (Secs.~\ref{subsec:mpc}--\ref{subsec:wbc}).  %
\end{enumerate}

\looseness=-1 In horizontal slides on Unitree Go1 hardware (Sec.~\ref{sec:results}), SCRUB completed 5/5 flat-wall traverses, 5/5 orbits of a constant-curvature column, and 89 of 90 trials in a speed and load sweep on both surfaces. With each condition tuned individually (Sec.~\ref{subsec:protocol}), no orbit completed with a fixed normal, our reimplementation of the friction-compensated normal~\cite{nasiri2024hybrid}, either row removed, a velocity row without gait averaging, or the MPC's tool reaction removed. An MPC that plans only the normal part of the tool reaction completed 2/5 orbits.

%% file: sections/fig_architecture.tex
\begin{figure*}[!t]
  \centering
  \includegraphics[width=0.90\textwidth]{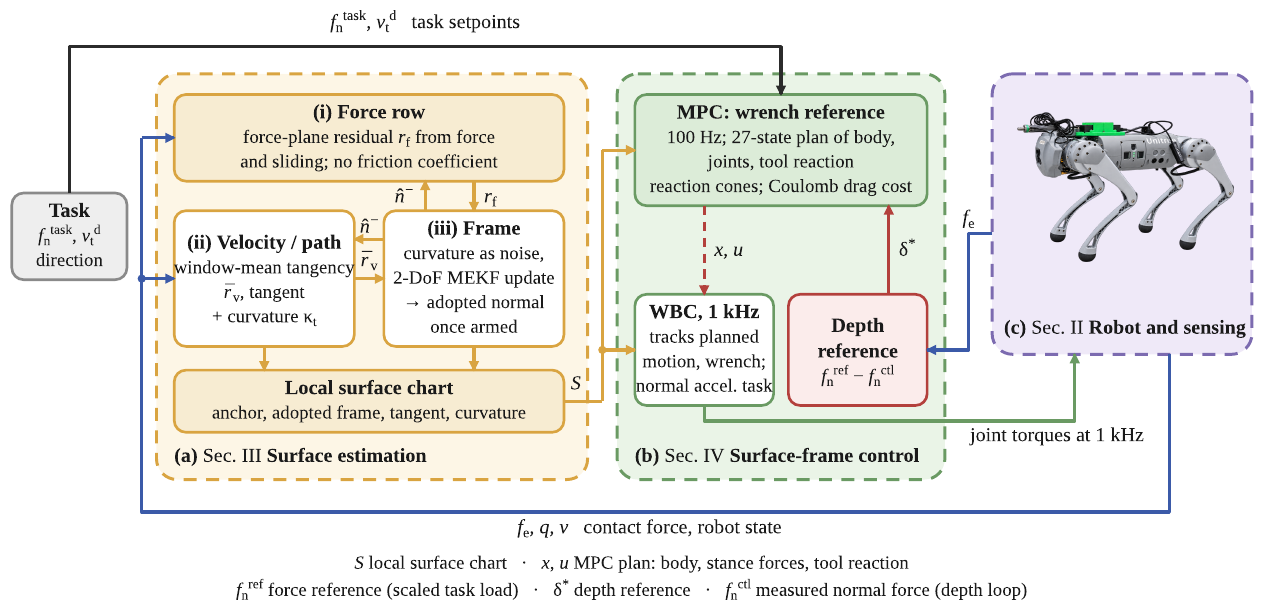}
  \caption{\textbf{SCRUB overview.}  (a) Estimation builds the chart $S$ from force and tool
  motion.  (b) MPC, depth reference, and WBC turn $S$ into joint torques
  for (c) the robot, which returns contact force and state.  Photo in (c):
  background removed.}
  \label{fig:architecture}
\end{figure*}

%% file: sections/03_setup.tex
\section{System, Problem Setup, and Assumptions}
\label{sec:setup}

\noindent\textbf{Task and sensing.}
Facing the surface, the robot walks sideways in a crawl gait, along the
wall or around the column.  With no map, it must slide the tool's tip
center at commanded speed $v_t^d$ and hold the \emph{task load}
$f_n^{\mathrm{task}}$, the target normal force.

\looseness=-1 An inline IIT FT17 force/torque sensor measures the contact force $f_e$~\cite{iit2018ftsensors}.  For odometry, a contact-aided invariant extended Kalman filter (EKF) fuses inertial and leg-kinematic data~\cite{hartley2020contact}.  Unless stated otherwise, we use the
gravity-aligned world frame ($z$ up).  We model the tool as rigid,
with nominal tip center $p_e$ at $r_e$ in the trunk frame.  The tool
Jacobian $J_e(q)$, with $q$ being the base and joint positions, maps the base and
joint velocities $v$ to the tip-center velocity $v_e=J_ev$.  As the tool shares
the trunk's angular velocity $\omega_e$, its spherical tip can roll, with
contact radius $r_c$ from the tip center to the contact point.
Tables~\ref{tab:notation} and~\ref{tab:params} list notation and settings.

\begin{table}[!h]
 \centering
 \caption{Notations used in this work.  A hat marks an estimate.}
 \label{tab:notation}
 \scriptsize
 \setlength{\tabcolsep}{3pt}
 \renewcommand{\arraystretch}{1.0}
 \begin{tabular}{@{}>{\raggedright\arraybackslash}p{0.16\columnwidth}
                  >{\raggedright\arraybackslash}p{0.80\columnwidth}@{}}
  \toprule
  \thd{Symbol} & \thd{Meaning} \\
  \midrule
  \grouprow{2}{Surface and frames}
  $n$ & true outward surface normal (toward the robot) \\
  $u_{\mathrm{slip}}$ & true sliding direction of the tip on the surface \\
  $\hat t$, $\hat b$ & slide tangent; binormal $\hat b=\hat n_s\times\hat t$ \\
  $\kappa_t$ & chart curvature along $\hat t$ (controller only) \\
  \grouprow{2}{Estimator}
  $\hat n^-$ & prior normal estimate (before an update) \\
  $\hat n^+$ & posterior normal estimate (after an update) \\
  $\hat n_s$ & adopted normal, used by the MPC and WBC \\
  $\hat n_f$ & force-direction normal (before arming; depth loop) \\
  $\hat u_m$ & estimated sliding direction \\
  $\hat m$ & estimated cross-slide axis, $\hat n^-\times\hat u_m$ \\
  $\theta_u$ & along-slide tilt from $\hat n^-$ to $n$ (toward $\hat u_m$) \\
  $\theta_m$ & cross-slide tilt from $\hat n^-$ to $n$ (toward $\hat m$) \\
  $v_e$, $v_c$ & tip-center velocity; contact-point velocity \\
  $r_f$ & force-row residual (force outside the estimated sliding plane) \\
  $\bar r_v$ & velocity-row residual (window mean of normal-velocity error) \\
  \grouprow{2}{Forces}
  $f_e$ & measured contact force \\
  $F_n$, $F_t$ & true load; true drag \\
  $f_n^m$ & measured force along $\hat n^-$ \\
  $f_n^{\mathrm{task}}$ & task load (target normal force) \\
  $f_n^{\mathrm{ref}}$ & force reference (task load, ramped in) \\
  $f_n^{\mathrm{ctl}}$ & measured force along $\hat n_f$ (depth loop) \\
  \grouprow{2}{Controller}
  $\delta$, $\delta^\star$ & depth; depth reference \\
  $v_n^c$ & commanded press rate, $-\dot\delta^\star$ (outward positive) \\
  $\lambda$ & planned tool reaction $[\lambda_n\ \lambda_t\ \lambda_b]$ in the MPC \\
  \bottomrule
 \end{tabular}
\end{table}

\begin{table}[!h]
 \centering
 \caption{SCRUB implementation settings.}
 \label{tab:params}
 \scriptsize
 \setlength{\tabcolsep}{3pt}
 \renewcommand{\arraystretch}{1.0}
 \begin{tabular}{@{}>{\raggedright\arraybackslash}p{0.47\columnwidth}
                  >{\raggedright\arraybackslash}p{0.17\columnwidth}
                  >{\raggedright\arraybackslash}p{0.31\columnwidth}@{}}
  \toprule
  \thd{Setting} & \thd{Symbol} & \thd{Value} \\
  \midrule
  \grouprow{3}{Estimator}
  Force-row noise & $\sigma_{f,0}$, $\beta_f$ & 0.3\,N, 0.05 \\
  Friction bound & $\mu_{\max}$ & 0.8 \\
  Stiction bound (start) & $\mu_s$ & 0.6 \\ %
  Velocity noise & $\sigma_{v,0}$ & 2\,mm/s \\
  Load--depth slope, lower bound & $k_{\min}$ & 200\,N/m \\
  Velocity-row window & $T_w$ & 0.5\,s \\
  Independent samples per window & $N_{\mathrm{eff}}$ & 30 \\
  Odometry drift & $v_{\mathrm{drift}}$ & 1\,mm/s \\ %
  Curvature bound; correlation length & $\kappa_{\max}$, $\ell_c$ & 10\,m$^{-1}$, 1\,mm \\
  Contact radius, stiffness & $r_c$, $k_c$ & 11\,mm, 800\,N/m \\
  \grouprow{3}{Planner and controller}
  Gait & -- & crawl, 0.5\,s cycle \\
  MPC rate / horizon & -- & 100\,Hz / 1.0\,s \\
  Reaction lag & $\tau_F$ & 0.05\,s \\
  MPC reaction weights & $w_n,w_t,w_b$ & 920, 100, 1000\,N$^{-2}$ \\
  MPC depth weight & $w_\delta$ & $2\times10^4$\,m$^{-2}$ \\
  Drag coefficient & $\mu_k$ & 0.24 \\
  Binormal cone cap & $\mu_b$ & 0.02 \\
  Depth trim rate & $k_\delta$ & 3\,mm/(N\,s) \\
  Control and estimator rate & -- & 1\,kHz \\
  \bottomrule
 \end{tabular}
\end{table}

\looseness=-1 \noindent\textbf{Local surface chart.}
The controller's only surface model is the \emph{chart} $S$ near the tip,
built in Sec.~\ref{sec:estimation}.  In the chart's frame
$(\hat t,\hat b,\hat n_s)$, the \emph{adopted normal} $\hat n_s$ points
outward, the tangent $\hat t$ runs along the
slide, and the binormal $\hat b=\hat n_s\times\hat t$ across it
(Fig.~\ref{fig:go1_axes}).  Along $\hat t$, the chart bends with
curvature $\kappa_t$.  The \emph{anchor} $\hat p_s$ starts at the tip
center at first contact.  The nominal contact point is
$p_c=p_e-r_c\hat n_s$.

\looseness=-1 The chart defines the depth $\delta=\hat n_f^\top(\hat p_s-p_e)$, the
distance the tip appears to press past the anchor along the
\emph{force-direction normal} $\hat n_f$ (the measured force direction with
the fitted drag along $\hat t$ removed, Sec.~\ref{subsec:fusion}).  Since
pressing mostly compresses the bumper, $p_e$ appears to sink past the surface.  Depth includes
this sinking but differs from the bumper's compression by an unknown
zero-load depth, which the depth loop of Sec.~\ref{subsec:ref} estimates online.

\looseness=-1 \noindent\textbf{Contact model.}
Given the true normal $n$ and sliding direction $u_{\mathrm{slip}}$, we
model $f_e=F_nn-F_tu_{\mathrm{slip}}+\eta_f$, with load $F_n>0$, drag
$F_t\ge0$, and noise $\eta_f$.

\begin{assumption}[Sliding contact]
\label{ass:contact}
The contact is compressive and sliding, with axisymmetric
pressure~\cite{johnson1985contact}.  Load rises with
depth, with slope at least $k_{\min}$ and no hysteresis.  Friction follows
Coulomb's law, proportional to load with the same coefficient in every
direction, and acts exactly opposite to the motion.  Drag therefore lies in
the sliding plane.  The noise $\eta_f$ includes surface texture, asymmetric
shear, and the uncompensated inertia of the tip beyond the sensor.
\end{assumption}

\begin{assumption}[Surface geometry]
\label{ass:geometry}
The surface is near vertical, with normal curvature bounded by
$\kappa_{\max}$ along the horizontal path.  Since the path neither curves
within the surface nor twists, we neglect geodesic curvature and torsion.
\end{assumption}

\begin{assumption}[Start]
\label{ass:start}
Each trial starts in contact, with the tool roughly
perpendicular to the surface.
\end{assumption}

\input{sections/fig_cues}

%% file: sections/fig_cues.tex
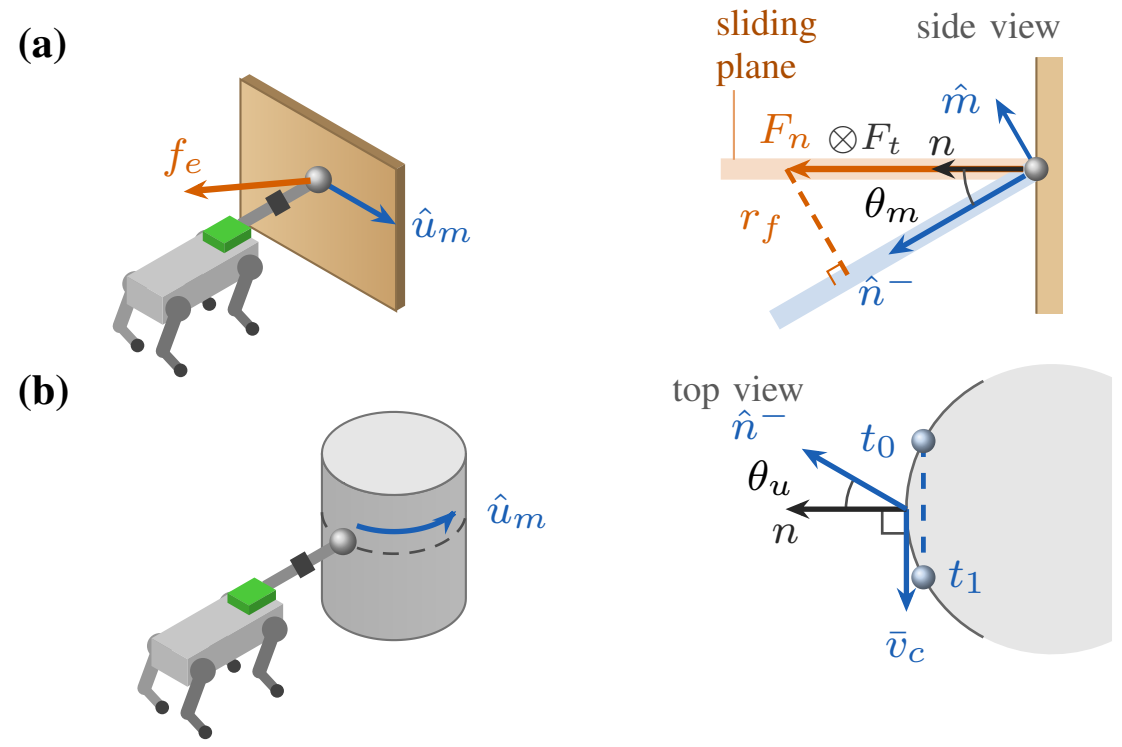
\begin{figure}[!t]
 \centering
 \definecolor{cueblue}{HTML}{1A5FB4}%
 \definecolor{cueorange}{HTML}{D55E00}%
 \definecolor{osbA}{HTML}{E2C394}%
 \definecolor{osbB}{HTML}{C9A06A}%
 \definecolor{botgreen}{HTML}{4CC93A}%
 \def\cueS{0.56}%
 \tikzset{iso/.style={x={({-0.7071*\cueS*1cm},{-0.4082*\cueS*1cm})},
                      y={({0.7071*\cueS*1cm},{-0.4082*\cueS*1cm})},
                      z={(0cm,{0.8165*\cueS*1cm})}}}
 \newcommand{\cueLeg}[3]{%
   \draw[#3, line width=2.6pt, line cap=round] (#1,#2,-0.34) -- (#1+0.26,#2,-0.82);
   \draw[#3, line width=1.8pt, line cap=round] (#1+0.26,#2,-0.82) -- (#1-0.04,#2,-1.25);
   \fill[black!75] (#1-0.04,#2,-1.25) circle (0.045cm);
   \fill[#3] (#1,#2,-0.34) circle (0.085cm);}
 \newcommand{\cueRobot}[1]{%
   \draw[black!45, line width=2.2pt] (#1+0.08,0,0) -- (#1+1.3,0,0);
   \draw[black!75, line width=4pt] (#1+0.55,0,0) -- (#1+0.8,0,0);
   \cueLeg{#1+2.7}{-0.36}{black!40}\cueLeg{#1+1.5}{-0.36}{black!40}%
   \fill[black!22] (#1+1.3,-0.3,-0.1) -- (#1+2.9,-0.3,-0.1) -- (#1+2.9,0.3,-0.1) -- (#1+1.3,0.3,-0.1) -- cycle;
   \fill[black!36] (#1+1.3,0.3,-0.1) -- (#1+2.9,0.3,-0.1) -- (#1+2.9,0.3,-0.5) -- (#1+1.3,0.3,-0.5) -- cycle;
   \fill[black!29] (#1+2.9,-0.3,-0.1) -- (#1+2.9,0.3,-0.1) -- (#1+2.9,0.3,-0.5) -- (#1+2.9,-0.3,-0.5) -- cycle;
   \fill[botgreen] (#1+1.3,-0.18,0.04) -- (#1+1.75,-0.18,0.04) -- (#1+1.75,0.18,0.04) -- (#1+1.3,0.18,0.04) -- cycle;
   \fill[botgreen!70!black] (#1+1.3,0.18,0.04) -- (#1+1.75,0.18,0.04) -- (#1+1.75,0.18,-0.1) -- (#1+1.3,0.18,-0.1) -- cycle;
   \fill[botgreen!80!black] (#1+1.75,-0.18,0.04) -- (#1+1.75,0.18,0.04) -- (#1+1.75,0.18,-0.1) -- (#1+1.75,-0.18,-0.1) -- cycle;
   \cueLeg{#1+1.5}{0.36}{black!55}\cueLeg{#1+2.7}{0.36}{black!55}%
   \shade[ball color=black!25] (#1,0,0) circle (0.09cm);}
 \begin{tikzpicture}[font=\footnotesize, line width=0.7pt,
   >={Stealth[length=4.5pt,width=3.6pt]}]
  \node[anchor=north west, font=\footnotesize\bfseries, inner sep=0pt] at (0,1.0) {(a)};
  \begin{scope}[shift={(2.0,-0.02)}, iso]
   \fill[osbB!80!black] (0,-1.3,0.8) -- (0,1.3,0.8) -- (-0.15,1.3,0.8) -- (-0.15,-1.3,0.8) -- cycle;
   \fill[osbB!65!black] (0,1.3,-1.25) -- (0,1.3,0.8) -- (-0.15,1.3,0.8) -- (-0.15,1.3,-1.25) -- cycle;
   \shade[left color=osbA, right color=osbB, draw=osbB!70!black, line width=0.4pt]
     (0,-1.3,-1.25) -- (0,1.3,-1.25) -- (0,1.3,0.8) -- (0,-1.3,0.8) -- cycle;
   \draw[->, cueblue, line width=1pt] (0,0.15,0) -- (0,1.3,0) node[right=-1pt] {$\hat u_m$};
   \cueRobot{0}
   \draw[->, cueorange, line width=1.2pt] (0.08,-0.05,0) -- (1.3,-0.95,0) node[above=-1pt] {$f_e$};
  \end{scope}
  \begin{scope}[shift={(6.75,0.05)}, x=0.87cm, y=0.87cm]
   \node[anchor=south east, font=\scriptsize, black!65, inner sep=0pt] at (0.2,0.98) {side view};
   \fill[osbA] (0,-1.1) rectangle (0.2,0.85);
   \draw[osbB!70!black, line width=0.4pt] (0,-1.1) -- (0,0.85);
   \draw[cueorange!22, line width=4pt] (0,0) -- (-2.4,0);
   \draw[cueorange!60, line width=0.4pt] (-2.3,0.07) -- (-2.3,0.6);
   \node[anchor=south west, font=\scriptsize, cueorange!80!black, align=left, inner sep=0pt] at (-2.45,0.62) {sliding\\plane};
   \draw[cueblue!22, line width=4pt] (0,0) -- ({-2.3*cos(30)},{-2.3*sin(30)});
   \draw[->, cueorange, line width=1.2pt] (-0.1,0) -- (-1.9,0) node[above=0pt] {$F_n$};
   \draw[->, black!85, line width=0.9pt] (-0.1,0) -- (-0.8,0);
   \node[black!85, above=-1pt] at (-0.72,0) {$n$};
   \node[black!80, anchor=south, font=\scriptsize, inner sep=1pt] at (-1.3,0.06) {$\otimes F_t$};
   \draw[->, cueblue, line width=1pt] ({-0.1*cos(30)},{-0.1*sin(30)}) -- ({-1.3*cos(30)},{-1.3*sin(30)}) node[below=-1pt] {$\hat n^-$};
   \draw[->, cueblue, line width=0.8pt] ({-0.08*sin(30)},{0.08*cos(30)}) -- ({-0.62*sin(30)},{0.62*cos(30)}) node[left=-1pt] {$\hat m$};
   \draw[black!70, line width=0.5pt] (-0.55,0) arc[start angle=180, end angle=210, radius=0.55];
   \node at ({-1.12*cos(14)},{-1.12*sin(14)}) {$\theta_m$};
   \draw[cueorange, densely dashed, line width=0.9pt] (-1.9,0) -- ({-1.9*cos(30)*cos(30)},{-1.9*cos(30)*sin(30)});
   \draw[cueorange, line width=0.4pt] ($({-1.9*cos(30)*cos(30)},{-1.9*cos(30)*sin(30)})+({-0.12*cos(30)},{-0.12*sin(30)})$)
     -- ++({-0.12*sin(30)},{0.12*cos(30)}) -- ++({0.12*cos(30)},{0.12*sin(30)});
   \node[cueorange, left=0pt] at (-1.76,-0.44) {$r_f$};
   \shade[ball color=black!25] (0,0) circle (0.08cm);
  \end{scope}
  \node[anchor=north west, font=\footnotesize\bfseries, inner sep=0pt] at (0,-1.3) {(b)};
  \begin{scope}[shift={(2.5,-2.22)}, iso]
   \def\cueR{0.85}%
   \draw[black!30, densely dotted, line width=0.4pt] plot[domain=135:315, samples=40] ({\cueR*cos(\x)}, {\cueR*sin(\x)}, 0);
   \shade[left color=black!32, middle color=black!4, right color=black!28, draw=black!55, line width=0.4pt]
     plot[domain=-45:135, samples=50] ({\cueR*cos(\x)}, {\cueR*sin(\x)}, -1.25)
     -- plot[domain=135:-45, samples=50] ({\cueR*cos(\x)}, {\cueR*sin(\x)}, 0.85) -- cycle;
   \fill[black!10, draw=black!55, line width=0.4pt]
     plot[domain=0:360, samples=72] ({\cueR*cos(\x)}, {\cueR*sin(\x)}, 0.85) -- cycle;
   \draw[black!70, densely dashed, line width=0.5pt] plot[domain=-45:135, samples=50] ({\cueR*cos(\x)}, {\cueR*sin(\x)}, 0);
   \draw[->, cueblue, line width=1pt] plot[domain=14:105, samples=30] ({\cueR*cos(\x)}, {\cueR*sin(\x)}, 0.32);
   \node[cueblue, right=2pt] at ({\cueR*cos(105)}, {\cueR*sin(105)}, 0.32) {$\hat u_m$};
   \cueRobot{\cueR}
  \end{scope}
  \begin{scope}[shift={(6.85,-2.2)}, x=0.8cm, y=0.8cm]
   \node[anchor=north west, font=\scriptsize, black!65, inner sep=0pt] at (-3.15,1.12) {top view};
   \fill[black!10] (0.5,{-sqrt(1.44-0.25)}) arc[start angle={-asin(sqrt(1.44-0.25)/1.2)}, end angle={-360+asin(sqrt(1.44-0.25)/1.2)}, radius=1.2] -- cycle;
   \draw[black!60, line width=0.5pt] ({1.2*cos(118)},{1.2*sin(118)}) arc[start angle=118, end angle=242, radius=1.2];
   \coordinate (Q0) at ({1.2*cos(152)},{1.2*sin(152)});
   \coordinate (Q1) at ({1.2*cos(208)},{1.2*sin(208)});
   \coordinate (QM) at (-1.2,0);
   \draw[cueblue, dashed, line width=0.9pt] (Q0) -- (Q1);
   \draw[black!70, line width=0.5pt] ($(QM)+({0.5*cos(180)},{0.5*sin(180)})$) arc[start angle=180, end angle=150, radius=0.5];
   \draw[black!70, line width=0.5pt] ($(QM)+({0.2*cos(180)},{0.2*sin(180)})$) -- ++({0.2*cos(270)},{0.2*sin(270)}) -- ++({-0.2*cos(180)},{-0.2*sin(180)});
   \node at ($(QM)+({1.15*cos(167)},{1.15*sin(167)})$) {$\theta_u$};
   \draw[->, cueblue, line width=1pt] (QM) -- ++({0.85*cos(270)},{0.85*sin(270)}) node[below=0pt] {$\bar v_c$};
   \draw[->, black!85, line width=1pt] (QM) -- ++({1.0*cos(180)},{1.0*sin(180)}) node[below=-1pt] {$n$};
   \draw[->, cueblue, line width=1pt] (QM) -- ++({1.0*cos(150)},{1.0*sin(150)}) node[above left=-2pt] {$\hat n^-$};
   \shade[ball color=cueblue!35] (Q0) circle (0.08cm);
   \shade[ball color=cueblue!35] (Q1) circle (0.08cm);
   \node[cueblue, left=1pt] at (Q0) {$t_0$};
   \node[cueblue, right=1pt] at (Q1) {$t_1$};
  \end{scope}
 \end{tikzpicture}
 \vspace{-0.6ex}
 \caption{\textbf{Estimator cues} (tilts exaggerated).  (a) Force row, tool sliding out of
 the page.  The true (orange) and estimated (blue) sliding
 planes are edge-on.  Load $F_n$ acts in the page and drag $F_t$ ($\otimes$)
 into it, so $f_e$ stays in the orange plane.  Its component perpendicular to the blue plane,
 the force-row residual $r_f=-\hat m^\top f_e$, thus measures cross-slide tilt $\theta_m$.  (b)~Velocity row, top view.
 From $t_0$ to $t_1$, the tool moves by $T_w\bar v_c$ (dashed).  Without press
 motion, this displacement is ideally perpendicular to $n$ at mid-window.
 The prior $\hat n^-$ deviates from this right angle by along-slide tilt
 $\theta_u$.}
 \label{fig:cues}
\end{figure}

%% file: sections/04_estimation.tex
\section{Surface-Frame Estimation}
\label{sec:estimation}

\looseness=-1 A Kalman filter, run every control cycle, tracks the
surface normal from two measurement rows, force and velocity.  Its state is the small tilt $\theta$ from the
estimated to the true normal.  For small angles, $n\approx\hat n^-+E^-\theta$, where the prior $\hat n^-$ is
the current estimate and $E^-$ holds two directions perpendicular to it.  Each
update estimates $\theta$, rotates $\hat n^-$ by it into the posterior
$\hat n^+$, and resets $\theta$ to zero.  This filter is a multiplicative EKF (MEKF) with covariance $P$~\cite{markley2003attitude}.

\subsection{Force-Plane Residual}
\label{subsec:normal_estimation}

\looseness=-1 As friction tilts the force only within the true sliding plane
(Assumption~\ref{ass:contact}), the force row measures cross-slide tilt
(Fig.~\ref{fig:cues}(a)) from the force component perpendicular to the
estimated sliding plane of $\hat n^-$ and the estimated sliding direction
$\hat u_m$:
\begin{equation}
 \begin{aligned}
 v_c&=v_e-r_{\mathrm{eff}}(\omega_e\times\hat n^-),\qquad
 f_n^m=(\hat n^-)^\top f_e,\\
 s_m\hat u_m&=(I-\hat n^-(\hat n^-)^\top)v_c,\\
 \hat m&=\hat n^-\times\hat u_m,\qquad
 r_f=-\hat m^\top f_e ,
\end{aligned}
 \label{eq:force_plane}
\end{equation}
where the estimated contact-point velocity $v_c$ is the odometry
tip-center velocity $v_e$ minus the tip's rolling velocity, $s_m$ is the estimated
sliding speed, and $\theta=[\theta_u\ \theta_m]^\top$ in the basis
$E^-\approx[\hat u_m\ \hat m]$.  The rolling radius $r_{\mathrm{eff}}$ shrinks from $r_c$ as load grows and is zero above $r_ck_c=8.8\,\mathrm N$, with $k_c$ the contact stiffness.  

\begin{proposition}[Zero crossing independent of friction]
\label{prop:rf}
If the velocity used in~\eqref{eq:force_plane} lies in the plane of the true
normal and sliding direction, $v_c=s_tu_{\mathrm{slip}}+v_nn$ with sliding speed $s_t>0$ and
normal component $v_n$ (true if the bumper compresses only along $n$), then
\begin{equation}
 r_f=-A\,\hat m^\top n-\hat m^\top\eta_f,\quad
 A=F_n+F_t\frac{v_n}{s_t},\quad \hat m^\top n\approx\theta_m .
 \label{eq:force_plane_linearized}
\end{equation}
If $\hat m^\top\eta_f=0$ and $A\neq0$, then $r_f=0$ if and only if
$\hat m^\top n=0$, regardless of $F_t$.
\end{proposition}

\looseness=-1 \noindent The proposition follows from $v_c$
lying in the estimated sliding plane, which gives $\hat m\perp v_c$ and
$\hat m^\top u_{\mathrm{slip}}=-(v_n/s_t)\hat m^\top n$.  Friction
thus scales $A$ without moving the zero crossing, and noise adds only an
offset, leaving $A$ unchanged.  With no press motion, $v_n=0$ and
$v_c\parallel u_{\mathrm{slip}}$, so the drag has no component along $\hat m$
and $A=F_n$.  Otherwise, $A$ differs from $F_n$ by the fraction
$(F_t/F_n)\lvert v_n/s_t\rvert$ and stays positive for $v_n/s_t>-F_n/F_t$, so
the update still moves the estimate toward $n$.  Geometrically, $r_f=0$ if and only if $\hat n^-$ lies in the
plane of $v_c$ and $f_e$.  This condition needs no tangent velocity and is the first step of~\cite{focchi2018slip}.  A friction-compensated
normal~\cite{nasiri2024hybrid} also lies in this plane, and its friction fit
sets only the along-slide tilt, which the velocity row measures instead
(Sec.~\ref{subsec:chord}).  The force row therefore keeps only this plane
condition.  SCRUB still uses friction values
in the force row's noise variance, the initial covariance, $\hat n_f$, and the
MPC.

\looseness=-1 Proposition~\ref{prop:rf} assumes that $v_c$ lies in the true sliding plane. If the estimated sliding direction is off by an angle $\delta\phi$ within the surface, the drag, which opposes the true direction, gains a component along $\hat m$. The residual then becomes $r_f\approx-F_t\,\delta\phi$ even at zero tilt, and with the row's gain near $F_n$, this offset appears as a cross-slide tilt of about $(F_t/F_n)\,\delta\phi$. The sliding direction comes from odometry, whose drift speed $v_{\mathrm{drift}}$ rotates it by $\delta\phi\approx v_{\mathrm{drift}}/s_m$. With $F_t/F_n$ equal to the drag coefficient $\mu_k$, this false tilt is at most $1.7^\circ$ at $8\,\mathrm{mm/s}$ and $0.9^\circ$ at $16\,\mathrm{mm/s}$, and we treat it as velocity noise in the row's variance~\eqref{eq:measurement_weights}. A drag-to-load ratio above $\mu_k$ raises this tilt (Sec.~\ref{subsec:results_discussion}).

\looseness=-1 The force row's measurement matrix
$H_f=(\hat u_m\times f_e)^\top E^-$, linearized with $\hat u_m$ held fixed,
has a gain of about $f_n^m$ without the friction term of $A$.  With the friction bound $\mu_{\max}$, the row's noise variance instead bounds the drag that press motion, velocity noise, and rolling can add:
\begin{equation}
\begin{aligned}
 R_f&=\sigma_{f,0}^2+(\beta_ff_n^m)^2+\bigl(\mu_{\max}f_n^m/s_m\bigr)^2\,\Sigma_v,\\
 \Sigma_v&=(v_n^c)^2+\sigma_{v,0}^2+r_{\mathrm{eff}}^2\|\omega_e\|^2,
\end{aligned}
 \label{eq:measurement_weights}
\end{equation}
with force-noise terms $\sigma_{f,0}$ and $\beta_f$, and velocity noise
$\sigma_{v,0}$.  Since the drag term divides by $s_m$, the variance
grows at low sliding speed, where the conditions of
Proposition~\ref{prop:rf} may fail.

\input{sections/tab_params}
\subsection{Window-Mean Velocity Row}
\label{subsec:chord}

\looseness=-1 A tool sliding on a static surface moves along
it~\cite{karayiannidis2014online,focchi2018slip,mihaylova2001planar}. The
velocity row therefore measures along-slide tilt from $(\hat n^-)^\top v_c$.  On a
walking robot, two effects violate this tangency at each sample.  The depth
loop moves the tool at the commanded press rate $v_n^c$
(Sec.~\ref{subsec:ref}).  Gait ripple moves the tool at about
$\dot f_n^m/k$, with $k$ the load--depth slope, hence at most
$|\dot f_n^m|/k_{\min}$ (Assumption~\ref{ass:contact}).  Per-sample updates would not
average out gait ripple: each update multiplies the normal-velocity
residual by its sensitivity (the sliding velocity), and the gait modulates
both.  Prior update laws,
including one for periodic wiping, accumulate this product with a forgetting
factor~\cite{karayiannidis2014online,dimeas2020progressive}.

\looseness=-1 To remove press motion, we subtract $v_n^c$.  To average out gait
ripple, we average the residual and sensitivity separately over windows $T_w$
of one gait cycle:
\begin{equation}
 \begin{aligned}
 \bar r_v&=\overline{v_n^c-(\hat n^-)^\top v_c},\qquad
 \bar H_v=\overline{v_c^\top E^-},\\
 \bar R_v&=\overline{\sigma_{v,0}^2+(v_n^c)^2+(\dot f_n^m/k_{\min})^2}\,/N_{\mathrm{eff}}
 +v_{\mathrm{drift}}^2,
 \end{aligned}
\label{eq:tangency_residual}
\end{equation}
where overbars are window means and $N_{\mathrm{eff}}$ counts effectively
independent samples per window.  The remaining ripple enters only the
noise variance, through the low-pass filtered load rate $\dot f_n^m$.
Averaging does not shrink the measured odometry drift $v_{\mathrm{drift}}$,
which is correlated across windows.  Since each update carries $E^-$ along, the
window means and $P$ share coordinates.

\subsection{Fusion and Adopted Normal}
\label{subsec:fusion}

\looseness=-1 The force row mainly observes $\theta_m$ each control cycle, while the
velocity row mainly observes $\theta_u$ at each window's end.  Between updates, prediction keeps the estimate but grows $P$ with distance slid: $P\leftarrow P+q_\kappa\|\Delta p_\parallel\|I$, with $\Delta p_\parallel$ the tangential displacement per control cycle.  The term
$q_\kappa=\kappa_{\max}^2\ell_c$ treats curvature up to $\kappa_{\max}$ as
noise with correlation length $\ell_c$ (Assumption~\ref{ass:geometry}),
without a curvature state, which would feed the estimate's turning back into
the prediction.  Turning only about $\hat b$, the normal changes $\theta_u$
alone, and growing both tilt variances is conservative.  On a curve, the window-mean
velocity is tangent at mid-window (Fig.~\ref{fig:cues}(b)), half the window's
swept angle behind the tip, and the same growth covers this lag.

\looseness=-1 Odometry drift also enters the velocity residual.  On a slide in one
direction, only the sum of along-slide tilt and the drift angle
$\arctan(v_{\mathrm{drift}}/s_m)$ is observable.  After each velocity-row update, we therefore keep the
along-slide standard deviation of $P$ at or above this angle.  

\looseness=-1 At the first firm press, the filter starts with $\hat n^-=f_e/\|f_e\|$
(Assumption~\ref{ass:start}).
Stiction with coefficient up to $\mu_s$ can tilt the press
force, and the initial covariance $P_0=\arctan^2(\mu_s)I$ covers that tilt, as
in~\cite{karayiannidis2014online}.  The filter
\emph{arms} provisionally once its cross-slide standard deviation falls below
$15^\circ$ after $0.1\,\mathrm s$ of sliding, confirms arming after a total of
$1\,\mathrm s$ below $3^\circ$, and disarms after $1\,\mathrm s$ above
$10^\circ$.  Provisional arming switches the adopted normal $\hat n_s$ of the MPC and WBC
from $\hat n_f$ to $\hat n^+$.  The depth loop keeps $\hat n_f$, which we form by
subtracting the drag along $\hat t$ with a drag-to-load ratio fitted online.
Besides the arming thresholds, step limits, rate limits, and other confidence
gates bound estimates and commands.

\subsection{Path Fit and Curvature}
\label{subsec:curvature_estimation}

\looseness=-1 The tangent $\hat t$ is the commanded slide direction projected perpendicular
to $\hat n_s$.  The anchor $\hat p_s$ follows a least-squares quadratic fit to recent
tip-center positions, each shifted outward along $\hat n_f$ by its depth, in
signed arc length $\xi\le0$ from the newest point.  The
curvature $\kappa_t$, positive for a convex surface, is the inverse-variance
average of two estimates independent of the filter: the fit's second
derivative and an EKF on the path's Darboux frame that models curvature as a
random walk in arc length (Fig.~\ref{fig:hw_csv_traces}(e), (j)).  Before use, we filter $\kappa_t$ and shrink it
toward zero, more strongly when its variance is larger.

%% file: sections/05_control.tex
\section{Surface-Frame Control}
\label{sec:control}

\looseness=-1 The controller in Fig.~\ref{fig:architecture}(b) must press along
the adopted normal with the task load, slide the tip along the tangent at
the commanded speed, and keep balance while stepping.  It commands body
motion and press depth, not contact force.  Because the tip sits
$0.35\,\mathrm m$ ahead of the trunk center, balance requires the stance
feet to resist the push and the twist from drag.  

\subsection{Predictive Wrench Planning}
\label{subsec:mpc}

\looseness=-1 Each MPC update solves a centroidal optimal control problem with gait, base, and foot-force costs, as in~\cite{farshidian2020ocs2,leggedcontrol,grandia2023perceptive}.  Its relaxed-barrier friction cones discourage stance feet from sliding.  To let the MPC place the body and set foot
forces before the push, we add the tool reaction $\lambda$ as a state and its
command $u_\lambda$ as an input.  The MPC has 27 states (base pose, centroidal momentum,
12 joint angles, and $\lambda$) and 27 inputs (foot forces, joint
velocities, and $u_\lambda$).

\looseness=-1 The reaction $\lambda=[\lambda_n\ \lambda_t\ \lambda_b]^\top$ acts at the
contact point $p_c$ as the force
$\lambda_n\hat n_s+\lambda_t\hat t+\lambda_b\hat b$, ignoring
contact-patch moments.  Since the MPC does not couple $\lambda_n$ to $\delta$, the planned $\lambda$ is a target reaction, and the depth loop of Sec.~\ref{subsec:ref} sets the actual load.  A supervisor switches among acquisition, recovery, sliding, and
detachment, and the reaction is active only in the supervisor's contact modes.  Within each
plan, the reaction directions stay fixed, while the chart reference
rotates about $\hat b$ by $\kappa_t$ times the predicted distance along
$\hat t$.  This rotation stays small on the column.  Finally, we add a
first-order lag $\tau_F$, reaction cones $|\lambda_t|\le\mu_k\lambda_n$
and $|\lambda_b|\le\mu_b\lambda_n$, and quadratic costs:
\begin{equation}
\begin{aligned}
 &\dot\lambda=(u_\lambda-\lambda)/\tau_F,\\
 &\ell=\tfrac12\bigl[w_n(\lambda_n-f_n^{\mathrm{ref}})^2
 +w_t\bigl(\lambda_t+\mu_k\,\mathrm{sgn}(v_t)\,\lambda_n\bigr)^2\\
 &\qquad+w_b\lambda_b^2+w_\delta(\delta-\delta^\star)^2\bigr],
\end{aligned}
\label{eq:mpc_additions}
\end{equation}
with weights $w_n,w_t,w_b,w_\delta$, nominal drag coefficient $\mu_k$,
binormal cap $\mu_b$ (Table~\ref{tab:params}), and measured tool velocity
$v_t$ along $\hat t$.

\looseness=-1 In~\eqref{eq:mpc_additions}, the lag limits how fast the planned
reaction changes.  The depth and load costs track the depth reference
$\delta^\star$ (Sec.~\ref{subsec:ref}) and $f_n^{\mathrm{ref}}$, the task load
ramped in at contact.  The drag cost pulls $\lambda_t$ toward Coulomb drag
$\mu_k\lambda_n$, which opposes $v_t$, with $\mathrm{sgn}$ smoothed near $v_t=0$.  With
drag as a cost, $\lambda_t$ can leave the cone's edge and stay
inside the cone when the tool stops.
Like planners for sliding contacts~\cite{carius2019trajectory}, the MPC
needs a friction coefficient, and a wrong $\mu_k$ changes the preferred
drag and the tangential cone.  On the wall, for example, measured drag
exceeded the $\mu_k$ cone (Sec.~\ref{subsec:results_discussion}).
Sec.~\ref{subsec:baselines} tests pinning $\lambda_t$ and
$\lambda_b$ or removing the whole reaction.

\subsection{Path and Depth References}
\label{subsec:ref}

\looseness=-1 The path reference advances along $\hat t$ so that the tip center moves at
$v_t^d$, offset by the contact radius.  Since pressing deeper raises the
force (Assumption~\ref{ass:contact}), force and motion along the normal
cannot be set independently~\cite{mason1981compliance}.  We therefore
regulate force mainly by moving $\delta^\star$.  In contact, the depth loop
moves $\delta^\star$ toward the depth a spring model predicts for
$f_n^{\mathrm{ref}}$, plus a trim $\delta_I$ that integrates the force
error~\cite{villani2016force}:
\begin{equation}
 \delta^\star\to\hat\delta_0+f_n^{\mathrm{ref}}/\hat k+\delta_I,\qquad
 \dot\delta_I=k_\delta\,\bar e,
 \label{eq:depth_law}
\end{equation}
with $\hat\delta_0$ the zero-load depth, $\hat k$ the load--depth slope
fitted online by recursive least squares, and $k_\delta$ the trim rate.
The force error $\bar e$ is the low-pass filtered
$f_n^{\mathrm{ref}}-f_n^{\mathrm{ctl}}$, with $f_n^{\mathrm{ctl}}$ measured
along $\hat n_f$.  When the spring model mispredicts the load, the depth
loop corrects the zero-load depth toward $\delta-f_n^{\mathrm{ctl}}/\hat k$.
The commanded press rate $v_n^c=-\dot\delta^\star$ feeds the filter
(Sec.~\ref{subsec:chord}).

\subsection{Whole-Body Tracking}
\label{subsec:wbc}

\looseness=-1 The WBC tracks the interpolated MPC plan at the control rate with an inverse-dynamics quadratic program over accelerations, foot forces, and torques~\cite{zapolsky2014inverse}.  It minimizes weighted objectives
under hard constraints on the floating-base dynamics, stance-foot motion,
torques, and friction.  We feed the planned reaction forward as a
known external wrench and add
a weighted objective on the tool's normal acceleration,
$\hat n_s^\top\dot v_e=-\dot{\hat n}_s^\top v_e+a_F$, where $a_F$ is a small
correction from the force error.  With $a_F=0$ and ideal tracking, rotating
the adopted normal alone leaves the tool's normal velocity
$\hat n_s^\top v_e$ unchanged.  The depth reference therefore acts through
the planned motion, via the MPC's depth cost in~\eqref{eq:mpc_additions}.

%% file: sections/06_results.tex
\begin{figure*}[!t]
 \centering
 \includegraphics[width=0.42\textwidth]{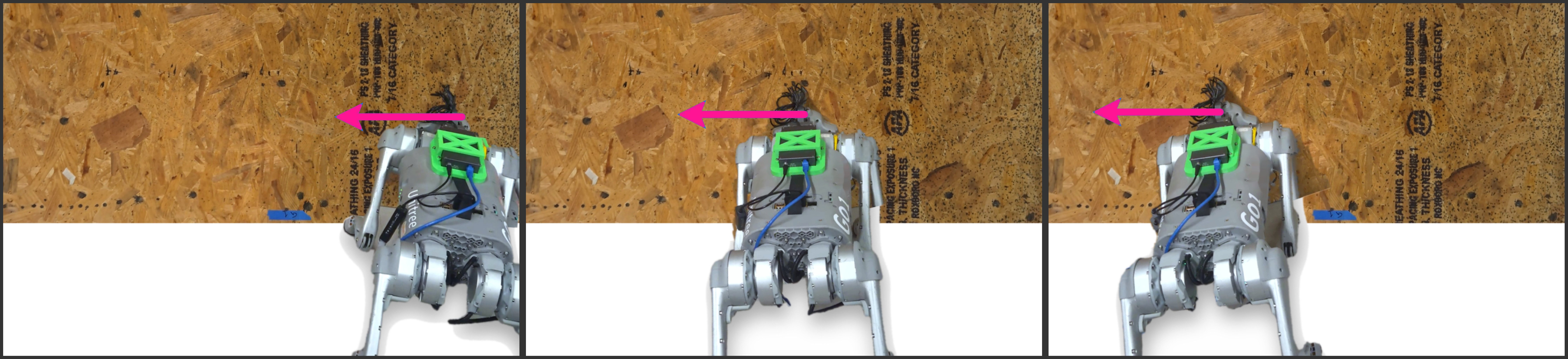}%
 \hspace{0.10\textwidth}%
 \includegraphics[width=0.42\textwidth]{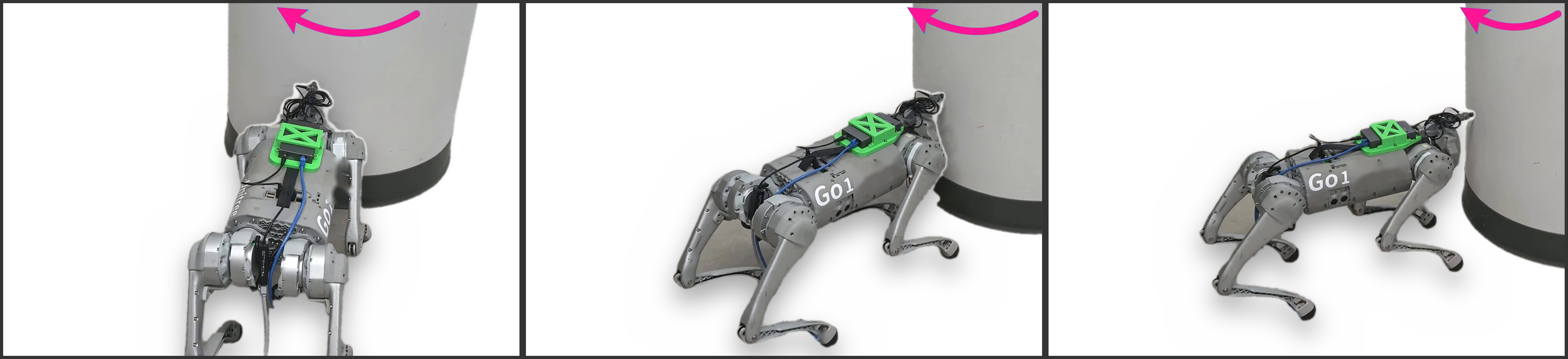}
 \vspace{-1.0ex}
 \caption{\textbf{Test surfaces.}  OSB wall (left) and drywall column (right).  Arrows: slide direction.  Backgrounds removed.}
 \label{fig:hw_strips}
 \vspace{-0.8ex}
\end{figure*}

\begin{figure*}[!t]
 \centering
 \begin{tikzpicture}
  \node[anchor=south west, inner sep=0pt] (traces)
   {\includegraphics[width=0.90\textwidth]{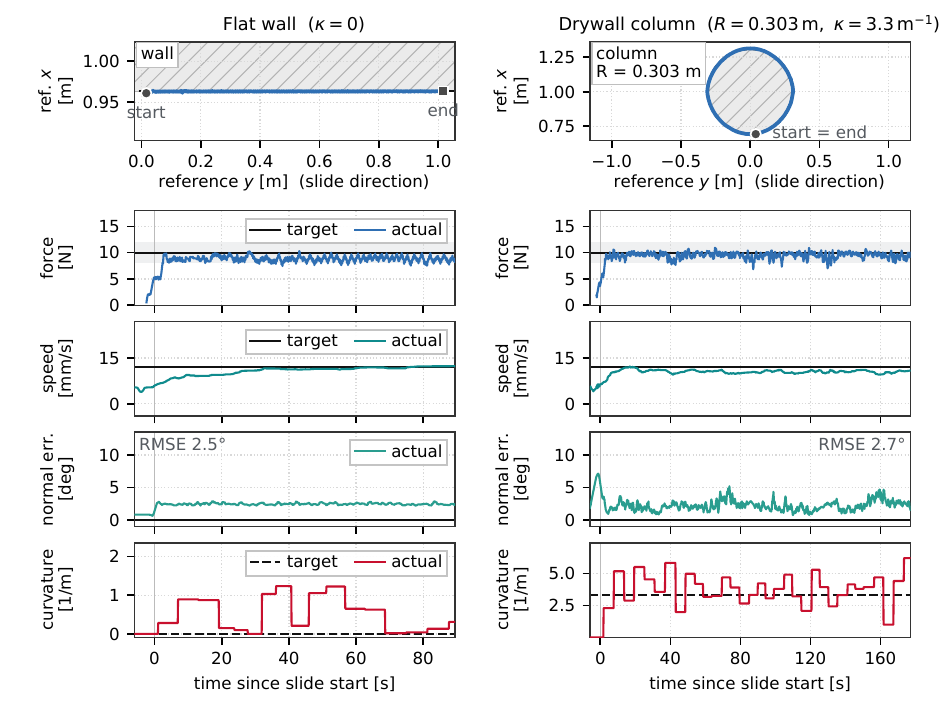}};
  \begin{scope}[x={(traces.south east)}, y={(traces.north west)},
                every node/.style={anchor=north west, inner sep=0pt, font=\footnotesize\sffamily\bfseries}]
   \foreach \y/\l/\r in {0.9409/a/f, 0.7009/b/g, 0.5434/c/h, 0.3853/d/i, 0.2279/e/j} {
    \node at (0.003, \y) {(\l)};
    \node at (0.494, \y) {(\r)};
   }
  \end{scope}
 \end{tikzpicture}
 \vspace{-1.0ex}
 \caption{\textbf{Example trials.}  Wall (a)--(e) and column (f)--(j) trials to completion.  (a), (f)~Tool path, with $x$ into the surface at the start and a dashed display-only fit.  (b), (g)~Force, with the $\pm2\,\mathrm N$ band shaded.  (c), (h)~Speed.  (d), (i)~Normal error, whose RMSE is for this trial.  (e), (j)~Curvature.  Black lines mark targets, dashed for the true curvature.  Force, speed, and normal error use a $0.1\,\mathrm s$ display window.}
 \label{fig:hw_csv_traces}
 \vspace{-0.8ex}
\end{figure*}
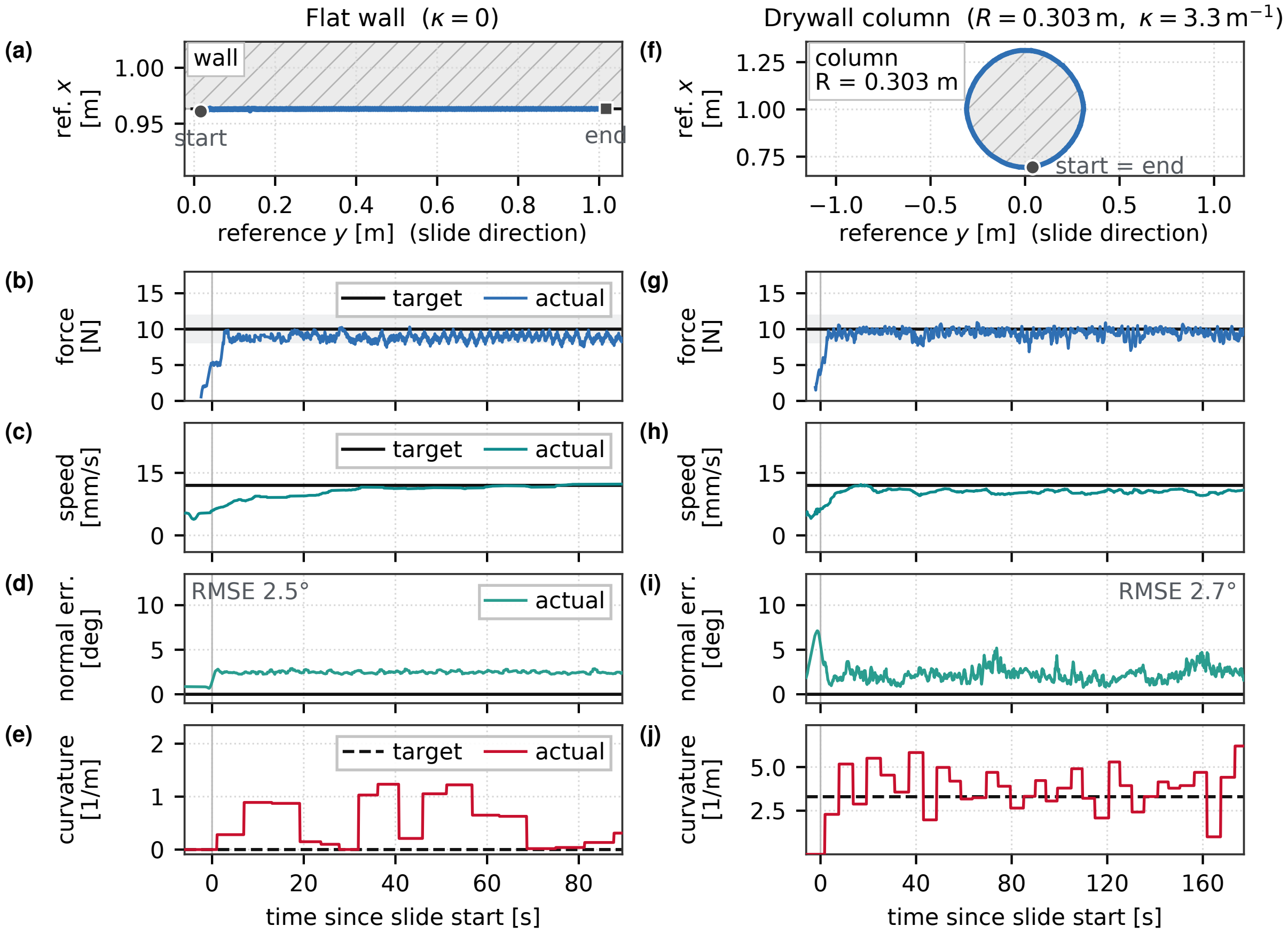

\section{Hardware Experiments and Discussion}
\label{sec:results}

This section evaluates completion, force and speed tracking, and normal accuracy on both surfaces and across speeds and loads. Column baselines and ablations then test which estimator parts an orbit needs and whether the controller needs the planned tool reaction.

\subsection{Setup, Protocol, and Scoring}
\label{subsec:protocol}
\label{subsec:platform}

\noindent\textbf{Platform and surfaces.}  A Unitree Go1 (stock mass
$12\,\mathrm{kg}$) carried a $50\,\mathrm{mm}$ PETG tool body, with the tip
centered at the nominal $r_e=[0.349\ 0\ 0]^\top\,\mathrm m$ in the trunk frame.  On a
tethered offboard PC, the MPC solved each $10\,\mathrm{ms}$ update in about
$8.3\,\mathrm{ms}$.  The robot scrubbed a flat oriented strand board (OSB)
wall whose top leaned $10^\circ$ away from the robot, and a vertical drywall
column of radius $0.303\,\mathrm m$ (Fig.~\ref{fig:hw_strips}).

\looseness=-1 \noindent\textbf{Runs.}   Before each trial, we zeroed the force/torque
sensor and hand-placed the tool, aiming for perpendicular contact.  Each
started at $70$--$100\%$ battery charge.  We commanded a horizontal slide at
$12\,\mathrm{mm/s}$ with a $10\,\mathrm N$ task load, and every condition
outside the speed and load sweep used this speed.  A trial was complete
after about $1\,\mathrm m$ on the wall or one full column orbit
(Fig.~\ref{fig:hw_csv_traces}(a), (f)).  A fall (trunk
roll above $0.20\,\mathrm{rad}$), a safety stop, or $20\,\mathrm s$ without
progress was a failure.  %

\looseness=-1 \noindent\textbf{Conditions.}   We tested SCRUB on both surfaces and
seven baselines and ablations on the column.  The first five change the estimator.
Fixed normal freezes the geometric normal of the approach plane at first
contact in the world frame (\mbox{$9.2$--$11.0^\circ$} from the reference
normal).  Friction-compensated normal replaces only SCRUB's force row.  Our
reimplementation covers only the friction compensation
of~\cite{nasiri2024hybrid}.  It fits a friction coefficient by least squares
and subtracts the Coulomb drag along $v_c$, whereas SCRUB subtracts it along
$\hat t$ to form $\hat n_f$.  The baseline's cross-slide measurement is the
$\hat b$ component of the normalized force left after its own drag
subtraction.  If the fitted coefficient is wrong, the leftover drag shifts
this measurement's zero unless sliding is along $\hat t$.  Instantaneous
velocity keeps the press-rate subtraction but applies the velocity row per
sample ($N_{\mathrm{eff}}=1$), like the instantaneous tangency
of~\cite{karayiannidis2014online,focchi2018slip}.  The No velocity row
ablation removes that row.  The No force row ablation zeroes the force-row
residual: the filter arms but keeps its initial cross-slide tilt.  The other
two reduce the MPC.  Scalar MPC keeps only $\lambda_n$ and its command,
without drag or binormal costs or cones.  Its dynamics fix the drag at
$\lambda_t=-\mu_k\,\mathrm{sgn}(v_t)\lambda_n$ and include no binormal
force.  The No tool force ablation removes $\lambda$ with its costs and cones
but keeps the depth cost.  Both keep the estimator, the depth loop, and the
WBC's normal-acceleration objective, but the WBC feeds forward the planned
reaction in Scalar MPC and the measured force $f_e$ in No tool force.  All
conditions shared placement and scoring.  We tuned each condition
individually, with SCRUB at the Table~\ref{tab:params} settings, and held
$\mu_k$ fixed after fitting it on earlier, unscored SCRUB column runs.  Each
condition then ran as one block of five trials with unchanged software,
starting with SCRUB.  We excluded no trials, and the filter confirmed arming in every trial.

\looseness=-1 \noindent\textbf{Scoring.}   We placed a tape-measured surface model to
touch the tip and face the tool's horizontal heading at the start.  At each
tool position logged by base odometry, the model gives the reference normal
in the world frame, radial from the column center or perpendicular to the
wall.  Odometry gives traverse and orbit as signed progress along the
slide.  Normal-angle error, whose root-mean-square error (RMSE) we report, is the
adopted normal's angle from the reference normal.  On these horizontal slides, the error along the force row's
cross-slide axis is the vertical-tilt error: the adopted normal's elevation
above horizontal minus the reference normal's.  For scoring only, we subtracted the wall's
inclinometer-measured lean from wall elevations to make the reference normal's
elevation $0^\circ$ on both surfaces.

\looseness=-1 \noindent\textbf{Metrics and statistics.}   Scoring uses $100\,\mathrm{Hz}$
samples from sliding onset until completion or termination, excluding
post-stop periods and recovery, which SCRUB's supervisor logs never show.
Raw force RMSE compares the unfiltered force along the reference normal with the
task load, and in band is the share of samples with this force within
$\pm2\,\mathrm N$ of the task load.  Low-load exposure is the time with
measured load below $0.4\,\mathrm N$, close to losing contact.  Speed RMSE
compares the commanded speed with the tip center's finite-difference
base-odometry speed along $\hat t$, median-filtered over a centered
$0.2\,\mathrm s$ window.
Since failed trials are shorter and per-metric medians may come from
different trials, we rank conditions by completion first.  Completion counts have
Clopper--Pearson $95\%$ intervals: $[0.48,1]$ for 5/5,
$[0.05,0.85]$ for 2/5, and $[0,0.52]$ for 0/5.

\subsection{Wall and Column Results}
\label{subsec:results_discussion}

\begin{table}[!t]
 \centering
 \caption{SCRUB results (medians of five trials).}
 \label{tab:hw_scrub}
 \scriptsize
 \setlength{\tabcolsep}{3pt}
 \renewcommand{\arraystretch}{1.0}
 \begin{tabular*}{\columnwidth}{@{\extracolsep{\fill}}l cc@{}}
  \toprule
  \thd{Metric} & \mc{\thd{OSB wall}}& \mc{\thd{Drywall column}}\\
  \midrule
  \grouprow{3}{Completion and coverage}
  Completed trials \unit{(of 5)} & 5 & 5\\
  Falls \unit{(of 5)} & 0 & 0\\
  Traverse \unit{[m]} & 1.000 & 1.904\\
  Orbit swept \unit{[deg]} & \mc{--}& 360.1\\
  \grouprow{3}{Surface and force tracking}
  Normal RMSE \unit{[deg]} & 2.29 & 2.48\\
  Force RMSE, raw \unit{[N]} & 2.97 & 2.06\\
  In band, $\pm2\,\mathrm N$ \unit{[\%]} & 63.2 & 81.2\\
  Low-load exposure \unit{[s]} & 0.00 & 0.00\\
  \grouprow{3}{Secondary diagnostics}
  Speed RMSE \unit{[mm/s]} & 4.52 & 2.66\\
  \grouprow{3}{Estimator internals}
  Posterior $\sigma$, smaller / larger axis \unit{[deg]} & \mc{0.83 / 5.84}&
  \mc{0.80 / 4.89}\\
  \bottomrule
 \end{tabular*}
\end{table}

\looseness=-1 \noindent\textbf{Completion, force, and speed.}   In Table~\ref{tab:hw_scrub}, SCRUB completed all ten trials without falls, although on both surfaces the planner used one $\mu_k=0.24$ fitted on the column and the median drag-to-load ratio was $0.53$ on OSB and $0.24$ on drywall.  Median raw force RMSE was $2.97\,\mathrm N$ on the wall and $2.06\,\mathrm N$ on the column.  Column progress was $11.0$--$12.5\,\mathrm{mm/s}$ across trials, against the commanded $12\,\mathrm{mm/s}$.  Fig.~\ref{fig:hw_csv_traces} shows one trial on each surface, with force in (b), (g), speed in (c), (h), and normal error in (d), (i).

\looseness=-1 \noindent\textbf{Normal accuracy.}  On the wall and column, median normal RMSE was $2.29^\circ$ and $2.48^\circ$, and median vertical-tilt RMSE was $2.09^\circ$ and $0.79^\circ$, respectively. In contrast, a horizontal normal, which assumes a vertical wall, would err by the wall's $10^\circ$ lean. The wall's larger vertical-tilt error exceeded the cross-slide posterior $\sigma$ of $0.83^\circ$ and may come from OSB's direction-dependent friction.

\begin{figure}[!t]
 \centering
 \includegraphics[width=\columnwidth]{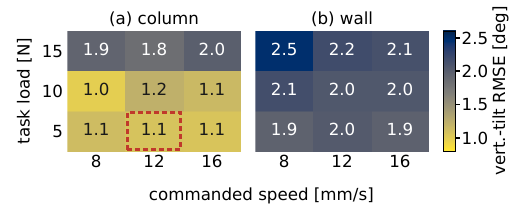}
 \vspace{-2.5ex}
 \caption{\textbf{Speed and load sweep.}  Median vertical-tilt RMSE of five trials per cell.  Dashed box: the cell with one fall.}
 \label{fig:sweep}
 \vspace{-1ex}
\end{figure}

\looseness=-1 \noindent\textbf{Speed and load sweep.}   We repeated both tasks five times at each pairing of $8$, $12$, and $16\,\mathrm{mm/s}$ with $5$, $10$, and $15\,\mathrm N$, with unchanged tuning (Fig.~\ref{fig:sweep}).  SCRUB completed $89$ of $90$ trials ($95\%$ interval $[0.94,1.00]$), and one column trial at $5\,\mathrm N$ fell.  Since the fifteen $5\,\mathrm N$ wall trials started further along the wall, the operator stopped them at its usable end, after about $0.95\,\mathrm m$, and we counted them complete.  Low-load exposure was $5$--$14\,\mathrm s$ at $5\,\mathrm N$ and at most $0.4\,\mathrm s$ otherwise.  Force RMSE was about $20\%$ of the task load on the column and $30\%$ on the wall.  Vertical-tilt RMSE changed little with speed but rose on the column to about $1.9^\circ$ at $15\,\mathrm N$.  At $12\,\mathrm{mm/s}$ and $10\,\mathrm N$, normal and force RMSE were within $0.25^\circ$ and $0.02\,\mathrm N$ of Table~\ref{tab:hw_scrub}, and column vertical-tilt RMSE was $1.2^\circ$, not $0.79^\circ$.

\subsection{Column Baselines and Ablations}
\label{subsec:baselines}

\begin{table}[!t]
 \centering
 \caption{Column baselines and ablations (medians of five trials).}
 \label{tab:hw_baselines}
 \label{tab:hw_planner}
 \scriptsize
 \setlength{\tabcolsep}{1.2pt}
 \renewcommand{\arraystretch}{1.0}
 \begin{tabular*}{\columnwidth}{@{\extracolsep{\fill}}l cccccccc@{}}
  \toprule
  \thd{Condition} & \mc{\thd{Done}} & \mc{\thd{Falls}} & \mc{\thd{Orbit}} & \mc{\thd{Normal}} & \mc{\thd{Force}} & \mc{\thd{In band}} & \mc{\thd{Low load}} & \mc{\thd{Speed}}\\
  & \mc{\unit{of 5 $\uparrow$}} & \mc{\unit{of 5 $\downarrow$}} & \mc{\unit{deg $\uparrow$}} & \mc{\unit{deg $\downarrow$}} & \mc{\unit{N $\downarrow$}} & \mc{\unit{\% $\uparrow$}} & \mc{\unit{s $\downarrow$}} & \mc{\unit{mm/s $\downarrow$}}\\
  \midrule
  SCRUB & \best{5} & \best{0} & \best{360.1} & \best{2.48} & \best{2.06} & 81.2 & \best{0.00} & \best{2.66}\\
  \grouprow{9}{Estimator baselines and ablations}
  Fixed normal & 0 & \best{0} & 16.1 & 30.88 & 4.94 & 43.1 & 0.29 & 16.9\\
  Friction comp. & 0 & 5 & 33.0 & 25.67 & 4.13 & 57.5 & 0.52 & 18.7\\
  Inst. velocity & 0 & 2 & 214.0 & 4.56 & 2.93 & 75.3 & 0.35 & 10.1\\
  No velocity row & 0 & 5 & 24.5 & 6.76 & 2.34 & 78.1 & 0.45 & 12.0\\
  No force row & 0 & 4 & 260.2 & 6.45 & 2.29 & 79.9 & 0.51 & 11.7\\
  \grouprow{9}{Planner ablations}
  Scalar MPC & 2 & \best{0} & 272 & 5.39 & 2.07 & \best{81.9} & 0.05 & 12.9\\
  No tool force & 0 & \best{0} & 102.5 & 7.22 & 4.32 & 35.3 & 0.7 & 8.8\\
  \bottomrule
 \end{tabular*}
 \par\vspace{2pt}
 \parbox{\columnwidth}{\raggedright Done, Orbit, Low load: completed trials, orbit swept, low-load exposure; Normal, Force, Speed: RMSE.  $\uparrow$/$\downarrow$: higher/lower is better; bold: best.}
\end{table}

\looseness=-1 \noindent\textbf{Estimator baselines and ablations.}   In Table~\ref{tab:hw_baselines}, none of the five estimator baselines and ablations completed an orbit.  Friction-compensated normal and No velocity row fell in all five trials, and No force row in four.  Fixed normal's $30.88^\circ$ RMSE comes from its $9.2$--$11.0^\circ$ start error and frozen estimate on a rotating surface.  With a fixed normal, the tool slid at about $31.6\,\mathrm{mm/s}$ against the commanded $12\,\mathrm{mm/s}$, covering a path as long as $32.5$--$35.2^\circ$ of column arc in about $5.7\,\mathrm s$.  Yet the tool left the orbit path, so net progress was $16.1^\circ$.  Instantaneous velocity swept a median $214^\circ$ with two falls, and its $4.56^\circ$ normal RMSE nearly doubled SCRUB's.

\looseness=-1 \noindent\textbf{Planner ablations.}   Without the MPC's tool reaction, no orbit completed: the No tool force ablation swept a median $102.5^\circ$ without falls, then lost contact and made no further progress.  Scoring excluded the recovery that followed.  Scalar MPC completed $2/5$.  In both planner ablations, normal RMSE exceeded SCRUB's, with the estimator and its settings unchanged.  Both also change the WBC feedforward, so neither isolates the MPC's reaction model.

\looseness=-1 \noindent\textbf{Discussion.}   On these slides, the normal turned only about the near-vertical binormal, and the force row corrected a constant vertical tilt.  We took the speed and $\mu_k$ from SCRUB's runs and ran its block first, choices that may favor SCRUB.

%% file: sections/07_conclusion.tex
\section{Conclusion and Future Work}
\label{sec:conclusion}

\looseness=-1 With SCRUB, an armless quadruped estimated the surface normal from contact alone while pressing and sliding a trunk-mounted tool on two unmapped, near-vertical surfaces.  Its Kalman filter fuses a force cue, whose zero does not depend on the friction coefficient if friction opposes the motion, with a gait-averaged velocity cue, and its MPC plans the tool reaction.  On Unitree Go1 hardware at $12\,\mathrm{mm/s}$ and $10\,\mathrm N$, SCRUB completed 5/5 (100\%) traverses of about $1\,\mathrm m$ on an OSB wall leaning $10^\circ$ and 5/5 (100\%) orbits of a drywall column of radius $0.303\,\mathrm m$, without falls.  Median normal RMSE was $2.29^\circ$ on the wall and $2.48^\circ$ on the column, and median force RMSE was $2.97\,\mathrm N$ and $2.06\,\mathrm N$.  Across three speeds of \mbox{$8$--$16\,\mathrm{mm/s}$}, three loads of $5$--$15\,\mathrm N$, and both surfaces, SCRUB completed 89 of 90 trials (98.9\%), with one column fall at $5\,\mathrm N$.

\looseness=-1 On the column, five estimator variants each completed 0/5 (0\%) orbits, with median normal RMSE of $4.56$--$30.88^\circ$: a fixed normal, a friction-compensated normal, a velocity row without gait averaging, and either row removed.  Of two planner ablations, an MPC planning only the normal reaction completed 2/5 (40\%) orbits, and one without the tool reaction 0/5 (0\%).  Within SCRUB's closed loop, an updated normal, both cues, gait averaging, and the planned tool reaction with its WBC feedforward were each needed.

\looseness=-1 The estimator assumes independent samples and may overstate its accuracy.  It has no odometry-drift state, omits gait-ripple velocity from the force row's noise, and never checks that this row's gain stays positive.  It runs outside the MPC, whose reaction has no contact moments, and the tool model ignores bumper shear.  Future work will test faster gaits and speeds above $16\,\mathrm{mm/s}$, changing curvature, non-planar paths, and varied friction, and will add normal estimation in the MPC, contact moments, a visual surface preview, and joint impedance in the WBC.

%% file: main.bbl
\begin{thebibliography}{10}
\providecommand{\url}[1]{#1}
\csname url@samestyle\endcsname
\providecommand{\newblock}{\relax}
\providecommand{\bibinfo}[2]{#2}
\providecommand{\BIBentrySTDinterwordspacing}{\spaceskip=0pt\relax}
\providecommand{\BIBentryALTinterwordstretchfactor}{4}
\providecommand{\BIBentryALTinterwordspacing}{\spaceskip=\fontdimen2\font plus
\BIBentryALTinterwordstretchfactor\fontdimen3\font minus
  \fontdimen4\font\relax}
\providecommand{\BIBforeignlanguage}[2]{{%
\expandafter\ifx\csname l@#1\endcsname\relax
\typeout{** WARNING: IEEEtran.bst: No hyphenation pattern has been}%
\typeout{** loaded for the language `#1'. Using the pattern for}%
\typeout{** the default language instead.}%
\else
\language=\csname l@#1\endcsname
\fi
#2}}
\providecommand{\BIBdecl}{\relax}
\BIBdecl

\bibitem{raibert1981hybrid}
M.~H. Raibert and J.~J. Craig, ``Hybrid position/force control of
  manipulators,'' \emph{ASME J. Dyn. Syst., Meas., Control}, vol. 103, no.~2,
  pp. 126--133, 1981.

\bibitem{khatib1987unified}
O.~Khatib, ``A unified approach for motion and force control of robot
  manipulators: The operational space formulation,'' \emph{IEEE J. Robot.
  Autom.}, vol.~3, no.~1, pp. 43--53, 1987.

\bibitem{mason1981compliance}
M.~T. Mason, ``Compliance and force control for computer controlled
  manipulators,'' \emph{IEEE Trans. Syst., Man, Cybern.}, vol. SMC-11, no.~6,
  pp. 418--432, 1981.

\bibitem{villani2016force}
L.~Villani and J.~De~Schutter, ``Force control,'' in \emph{Springer Handbook of
  Robotics}, 2nd~ed., B.~Siciliano and O.~Khatib, Eds.\hskip 1em plus 0.5em
  minus 0.4em\relax Cham, Switzerland: Springer, 2016, pp. 195--220.

\bibitem{orin2013centroidal}
D.~E. Orin, A.~Goswami, and S.-H. Lee, ``Centroidal dynamics of a humanoid
  robot,'' \emph{Auton. Robots}, vol.~35, no. 2--3, pp. 161--176, 2013.

\bibitem{sleiman2021unified}
J.-P. Sleiman, F.~Farshidian, M.~V. Minniti, and M.~Hutter, ``A unified {MPC}
  framework for whole-body dynamic locomotion and manipulation,'' \emph{IEEE
  Robot. Autom. Lett.}, vol.~6, no.~3, pp. 4688--4695, 2021.

\bibitem{johnson1985contact}
K.~L. Johnson, \emph{Contact Mechanics}.\hskip 1em plus 0.5em minus 0.4em\relax
  Cambridge, U.K.: Cambridge University Press, 1985.

\bibitem{bicchi1993contact}
A.~Bicchi, J.~K. Salisbury, and D.~L. Brock, ``Contact sensing from force
  measurements,'' \emph{Int. J. Robot. Res.}, vol.~12, no.~3, pp. 249--262,
  1993.

\bibitem{deschutter2007constraint}
J.~De~Schutter, T.~De~Laet, J.~Rutgeerts, W.~Decr{\'e}, R.~Smits,
  E.~Aertbeli{\"e}n, K.~Claes, and H.~Bruyninckx, ``Constraint-based task
  specification and estimation for sensor-based robot systems in the presence
  of geometric uncertainty,'' \emph{Int. J. Robot. Res.}, vol.~26, no.~5, pp.
  433--455, 2007.

\bibitem{karayiannidis2009adaptive}
Y.~Karayiannidis and Z.~Doulgeri, ``Adaptive control of robot contact tasks
  with on-line learning of planar surfaces,'' \emph{Automatica}, vol.~45,
  no.~10, pp. 2374--2382, 2009.

\bibitem{lin2020contact}
H.-C. Lin and M.~N. Mistry, ``Contact surface estimation via haptic
  perception,'' in \emph{Proc. IEEE Int. Conf. Robot. Autom. (ICRA)}, 2020, pp.
  5087--5093.

\bibitem{karayiannidis2014online}
Y.~Karayiannidis, C.~Smith, F.~E. Vi{\~n}a, and D.~Kragic, ``Online contact
  point estimation for uncalibrated tool use,'' in \emph{Proc. IEEE Int. Conf.
  Robot. Autom. (ICRA)}, 2014, pp. 2488--2494.

\bibitem{focchi2018slip}
M.~Focchi, V.~Barasuol, M.~Frigerio, D.~G. Caldwell, and C.~Semini, ``Slip
  detection and recovery for quadruped robots,'' in \emph{Robotics Research},
  ser. Springer Proc. Adv. Robot., A.~Bicchi and W.~Burgard, Eds.\hskip 1em
  plus 0.5em minus 0.4em\relax Cham, Switzerland: Springer, 2018, vol.~3, pp.
  185--199.

\bibitem{nasiri2024hybrid}
E.~Nasiri and L.~Wang, ``Hybrid force motion control with estimated surface
  normal for manufacturing applications,'' in \emph{Proc. Int. Conf. Ubiquitous
  Robots (UR)}, 2024, pp. 125--132.

\bibitem{portela2024force}
T.~Portela, G.~B. Margolis, Y.~Ji, and P.~Agrawal, ``Learning force control for
  legged manipulation,'' in \emph{Proc. IEEE Int. Conf. Robot. Autom. (ICRA)},
  2024, pp. 15\,366--15\,372.

\bibitem{bellicoso2019alma}
C.~D. Bellicoso, K.~Kr{\"a}mer, M.~St{\"a}uble, D.~V. Sako, F.~Jenelten,
  M.~Bjelonic, and M.~Hutter, ``{ALMA}: Articulated locomotion and manipulation
  for a torque-controllable robot,'' in \emph{Proc. IEEE Int. Conf. Robot.
  Autom. (ICRA)}, 2019, pp. 8477--8483.

\bibitem{arm2024pedipulate}
P.~Arm, M.~Mittal, H.~Kolvenbach, and M.~Hutter, ``Pedipulate: Enabling
  manipulation skills using a quadruped robot's leg,'' in \emph{Proc. IEEE Int.
  Conf. Robot. Autom. (ICRA)}, 2024, pp. 5717--5723.

\bibitem{sleiman2023multicontact}
J.-P. Sleiman, F.~Farshidian, and M.~Hutter, ``Versatile multicontact planning
  and control for legged loco-manipulation,'' \emph{Sci. Robot.}, vol.~8,
  no.~81, p. eadg5014, 2023.

\bibitem{rigo2023contact}
A.~Rigo, Y.~Chen, S.~K. Gupta, and Q.~Nguyen, ``Contact optimization for
  non-prehensile loco-manipulation via hierarchical model predictive control,''
  in \emph{Proc. IEEE Int. Conf. Robot. Autom. (ICRA)}, 2023, pp. 9945--9951.

\bibitem{mahmud2026contact}
A.~J. Mahmud, S.~Li, and X.~Wang, ``Contact as a decision variable:
  Capability-tradeoff contact selection for legged loco-manipulation,''
  arXiv:2609.30140 [cs.RO], 2026.

\bibitem{zeng2026force}
X.~Zeng, J.~Wang, T.~Zhang, L.~Zhang, B.~Xu, W.~Xia, Z.~Li, and Y.-H. Liu,
  ``Force-aware reinforcement learning with hybrid sensorless force estimation
  for wheeled-legged loco-manipulation,'' arXiv:2609.13779 [cs.RO], 2026.

\bibitem{samadi2021humanoid}
S.~Samadi, J.~Roux, A.~Tanguy, S.~Caron, and A.~Kheddar, ``Humanoid control
  under interchangeable fixed and sliding unilateral contacts,'' \emph{IEEE
  Robot. Autom. Lett.}, vol.~6, no.~2, pp. 4032--4039, 2021.

\bibitem{shi2026minimalist}
H.~Shi, S.~Hu, Y.~Hou, W.~Wang, C.~K. Liu, and S.~Song, ``Minimalist compliance
  control,'' in \emph{Proc. Robot.: Sci. Syst. (RSS)}, 2026.

\bibitem{farshidian2020ocs2}
F.~Farshidian \emph{et~al.}, ``{OCS2}: An open source library for optimal
  control of switched systems,'' GitHub repository, 2020, [Online]. Available:
  \url{https://github.com/leggedrobotics/ocs2}.

\bibitem{leggedcontrol}
Q.~Liao \emph{et~al.}, ``{legged\_control}: {NMPC}, {WBC}, state estimation,
  and sim2real framework for legged robots based on {OCS2} and ros-controls,''
  GitHub repository, 2022, [Online]. Available:
  \url{https://github.com/qiayuanl/legged_control}.

\bibitem{zapolsky2014inverse}
S.~Zapolsky and E.~Drumwright, ``Quadratic programming-based inverse dynamics
  control for legged robots with sticking and slipping frictional contacts,''
  in \emph{Proc. IEEE/RSJ Int. Conf. Intell. Robots Syst. (IROS)}, 2014, pp.
  3266--3271.

\bibitem{minniti2021model}
M.~V. Minniti, R.~Grandia, K.~F{\"a}h, F.~Farshidian, and M.~Hutter, ``Model
  predictive robot-environment interaction control for mobile manipulation
  tasks,'' in \emph{Proc. IEEE Int. Conf. Robot. Autom. (ICRA)}, 2021, pp.
  1651--1657.

\bibitem{iit2018ftsensors}
{Istituto Italiano di Tecnologia}, ``{FT-17} 6-axis force/torque sensor
  datasheet,'' Genoa, Italy, 2018.

\bibitem{hartley2020contact}
R.~Hartley, M.~Ghaffari, R.~M. Eustice, and J.~W. Grizzle, ``Contact-aided
  invariant extended {K}alman filtering for robot state estimation,''
  \emph{Int. J. Robot. Res.}, vol.~39, no.~4, pp. 402--430, 2020.

\bibitem{markley2003attitude}
F.~L. Markley, ``Attitude error representations for {K}alman filtering,''
  \emph{J. Guid. Control Dyn.}, vol.~26, no.~2, pp. 311--317, 2003.

\bibitem{mihaylova2001planar}
L.~Mihaylova, H.~Bruyninckx, J.~De~Schutter, and E.~Staffetti, ``Planar contour
  tracking in the presence of pose and model errors by {K}alman filtering
  techniques,'' in \emph{Proc. Int. Conf. Multisensor Fusion Integr. Intell.
  Syst. (MFI)}, 2001, pp. 329--334.

\bibitem{dimeas2020progressive}
F.~Dimeas and Z.~Doulgeri, ``Progressive automation of periodic tasks on planar
  surfaces of unknown pose with hybrid force/position control,'' in \emph{Proc.
  IEEE/RSJ Int. Conf. Intell. Robots Syst. (IROS)}, 2020, pp. 5246--5252.

\bibitem{grandia2023perceptive}
R.~Grandia, F.~Jenelten, S.~Yang, F.~Farshidian, and M.~Hutter, ``Perceptive
  locomotion through nonlinear model-predictive control,'' \emph{IEEE Trans.
  Robot.}, vol.~39, no.~5, pp. 3402--3421, 2023.

\bibitem{carius2019trajectory}
J.~Carius, R.~Ranftl, V.~Koltun, and M.~Hutter, ``Trajectory optimization for
  legged robots with slipping motions,'' \emph{IEEE Robot. Autom. Lett.},
  vol.~4, no.~3, pp. 3013--3020, 2019.

\end{thebibliography}
